\documentclass[letterpaper, 10 pt, conference]{ieeeconf}  

\IEEEoverridecommandlockouts                              

\usepackage{graphics} 
\usepackage{epsfig} 
\usepackage{mathptmx} 
\usepackage{times} 
\usepackage{amsmath} 
\usepackage{amssymb}  
\usepackage{subcaption}
\usepackage{textcomp}
\usepackage{stfloats}
\usepackage{url}
\usepackage{verbatim}
\usepackage{graphicx}
\usepackage{cite}
\usepackage{hyperref}
\usepackage{bbm}
\usepackage{bm}
\usepackage{color}
\usepackage{algorithm}
\usepackage{algpseudocode}
\usepackage{dsfont}
\algnewcommand\algorithmicinput{\textbf{\quad Input:}}
\algnewcommand\INPUT{\item[\algorithmicinput]}
\algnewcommand\algorithmicfind{\qquad Find}
\algnewcommand\FIND{\item[\algorithmicfind]}
\algnewcommand\algorithmicdefine{\qquad Define}
\algnewcommand\DEFINE{\item[\algorithmicdefine]}

\def\BibTeX{{\rm B\kern-.05em{\sc i\kern-.025em b}\kern-.08em
    T\kern-.1667em\lower.7ex\hbox{E}\kern-.125emX}}
\usepackage{balance}

\newcommand{\mR}{{\mathbb R}}

\newcommand{\mP}{{\mathbb P}}

\newcommand{\bG}{{\mathbf G}}

\newcommand{\bs}{{\mathbf s}}

\newcommand{\bk}{{\mathbf k}}

\newcommand{\by}{{\mathbf y}}

\newcommand{\bw}{{\mathbf w}}
\newcommand{\bx}{{\mathbf x}}
\newcommand{\be}{{\mathbf e}}
\newcommand{\bX}{{\mathbf X}}

\newcommand{\bu}{{\mathbf u}}

\newtheorem{theorem}{Theorem}

\newtheorem{lemma}{Lemma}
\newtheorem{assumption}{Assumption}
\newtheorem{remark}{Remark}

\newtheorem{problem}{Problem}

\title{\LARGE \bf
Safe Navigation using Density Functions
}

\author{ Andrew Zheng$^{*}$ and Sriram S.K.S Narayanan$^{*}$ and Umesh Vaidya
\thanks{*These authors contributed equally}
\thanks{Financial support from of NSF CPS award 1932458 and NSF
2031573 is greatly acknowledged. Andrew Zheng,  Sriram S.K.S Narayanan, and Umesh Vaidya are with the Department of Mechanical Engineering, Clemson University, Clemson, SC 29630, USA
        {\tt\small email: azheng@clemson.edu; sriramk@clemson.edu; uvaidya@clemson.edu}}%
}

\begin{document}

\maketitle
\thispagestyle{empty}
\pagestyle{empty}

\begin{abstract}

This paper presents a novel approach for safe control synthesis using the dual formulation of the navigation problem. The main contribution of this paper is in the analytical construction of density functions for almost everywhere navigation with safety constraints. In contrast to the existing approaches, where density functions are used for the analysis of navigation problems, we use density functions for the synthesis of safe controllers. We provide convergence proof using the proposed density functions for navigation with safety. Further, we use these density functions to design feedback controllers capable of navigating in cluttered environments and high-dimensional configuration spaces. The proposed analytical construction of density functions overcomes the problem associated with navigation functions, which are known to exist but challenging to construct, and potential functions, which suffer from local minima. Application of the developed framework is demonstrated on simple integrator dynamics and fully actuated robotic systems. Our project page with implementation is available at \url{https://github.com/clemson-dira/density_feedback_control}

\end{abstract}

\section{INTRODUCTION}

Safe navigation of mission-critical systems is of utmost importance in many modern autonomous applications. Autonomous vehicles and industrial robots are all critical applications in which there exists a need for navigation that adheres to safety constraints. Over the past decades, the general approach to the navigation problem has consisted of formulating compositions of the system that complies with the safety certification of the original system. This traditionally implies a hierarchical architecture that decomposes the navigation problem into planning and control \cite{lavalle2006planning}.

The planning problem involves defining a collision-free trajectory in the feasible configuration space given an initial and final configuration. These are typically implemented through sample-based planners such as rapidly-exploring random tree search (RRT) and probabilistic roadmaps (PRM) \cite{lavalle1998rapidly, amato1996randomized}. These sample-based methods are observed to be probabilistically complete through iterative samples of locally safe and feasible paths. Asymptotically optimal variations of these planners have been developed in \cite{karaman2011sampling}, where the convergence rate for optimality is improved in \cite{janson2015fast, gammell2015batch}.

Designing controllers to track these trajectories from the plan while satisfying dynamic and safety constraints is not so simple. Traditional methods, such as inverse dynamics, rely on the exact cancellation of the nonlinearities to track a resulting linear system through closed-loop control \cite{murray2017mathematical}. However, they do not guarantee safety in the presence of unsafe regions. More recently, control barrier functions (CBFs) have been introduced to provide safety certificates for the controller \cite{ames2019control}. However, CBFs only provide safety, so augmentation of CBFs with control Lyapunov functions (CLFs) is needed to guarantee convergence and safety \cite{ames2016control}.

This framework of hierarchical navigation has seen great success in many robotic applications \cite{di2018dynamic}; however, a natural issue of hierarchical navigation is the evaluation of safety certificates from the planning to the control level, which increases in complexity for large-scale systems \cite{singletary2022onboard}.

 \begin{figure}[t]
     \centering
     \includegraphics[width = \linewidth]{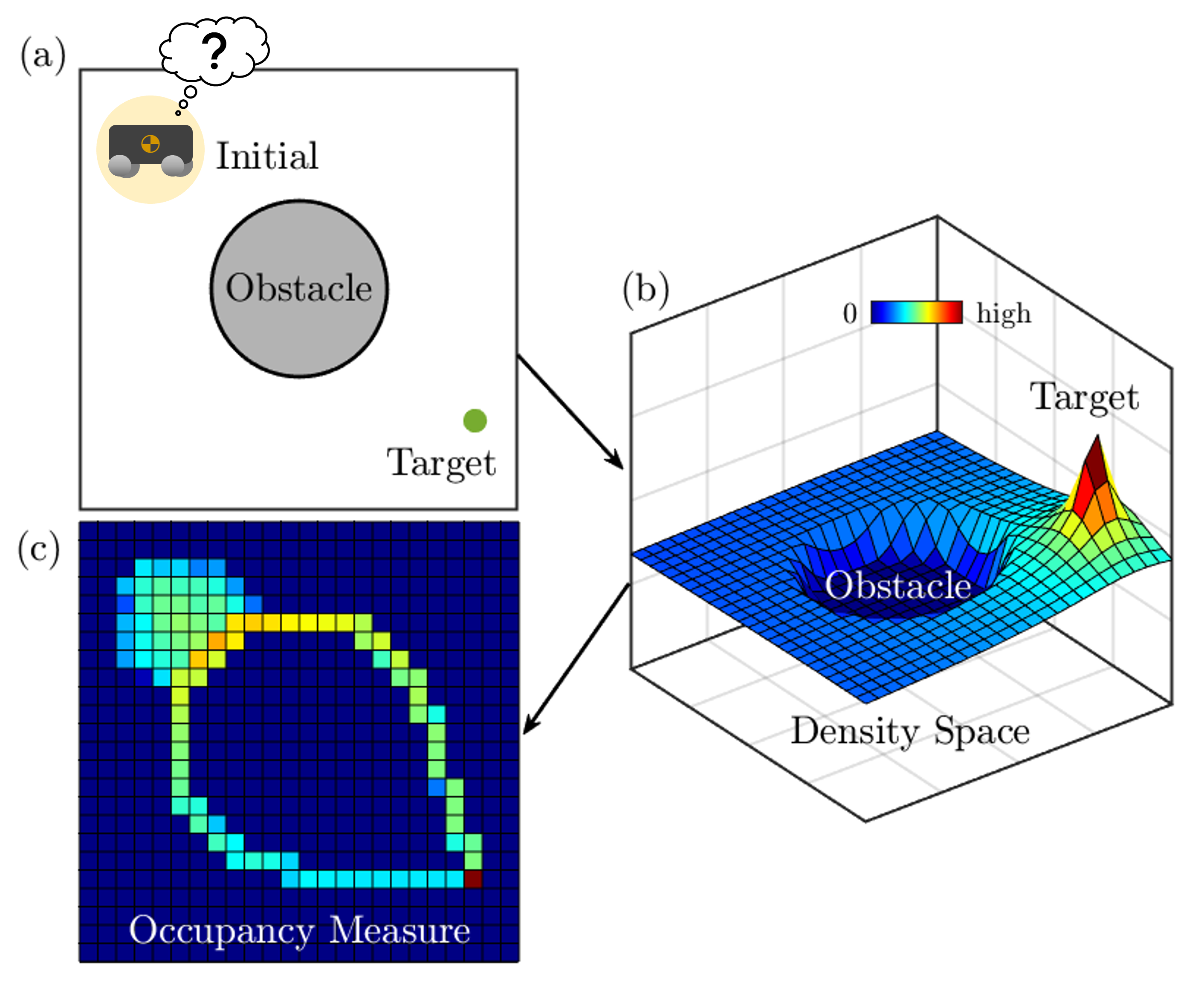}
     \caption{Navigation framework using density where (a) defines the navigation problem, (b) shows the density for navigation, and (c) shows occupancy measure, which physically denotes the duration of system trajectories occupying the set.}
     \label{fig:dens_nav_diagram}
 \end{figure}

A natural proposal is to jointly solve the navigation problem without the hierarchical structure. Artificial potential field based methods have attempted to solve the joint problem by the sum of attractive and repulsive potentials \cite{khatib1986real}. However, the existence of local minima is a well-known issue \cite{krogh1984generalized, rimon1990exact}. In \cite{rimon1990exact, KoditschekMotion}, a class of analytical potential functions, known as navigation functions (NFs), are introduced, which guarantees almost everywhere (a.e.) convergence while adhering to safety constraints. This method relies on a range of problem-specific tuning parameters to guarantee a.e. convergence. Moreso, complex safety constraints arising from arbitrarily shaped obstacles are limited by the possible mapping to a model sphere world.  Recent works have proposed using altered NF or conformal mapping to navigate complex unsafe sets \cite{loizou2011closed, filippidis2012navigation,fan2022robot}; however, these methods are nontrivial and physically unintuitive.

The navigation problem can alternatively be formulated in the dual space of density. In \cite{vaidya2018optimal}, a navigation measure was introduced to provide a convex formulation for synthesizing safe controllers. In the continuous-time setting, the density function was used as a safety certificate for the analysis and synthesis using the sum of squares optimization method \cite{rantzer2004analysis}. Similarly, density-based approaches are also used for the convergence analysis of existing navigation algorithms \cite{loizou2008density, dimarogonas2007application}. More recently, convex data-driven approaches based on the linear transfer Perron-Frobenius and Koopman operators are used for solving the optimal navigation problem with safety constraints \cite{yu2022data,10081458}. In contrast to using the convex dual formulation for navigation, we provide an analytical construction of density functions for navigation. In particular, the analytical construction of navigation density can be viewed as the dual construction of the classical NFs from \cite{rimon1990exact}. However, unlike \cite{rimon1990exact}, the construction is not restricted to navigation in the sphere world environment.

The main contribution of this paper is in providing analytical construction of density functions used for solving the safe navigation problem. The density function has a physical interpretation, where the measure associated with the density is a measure of occupancy of the system trajectories in any set of the state space as shown in Figure \ref{fig:dens_nav_diagram}. We exploit this occupancy-based physical interpretation of the density function in the construction of the navigation density functions. Unlike NFs, the density formulation can represent arbitrary shapes of the obstacle sets. We prove that the proposed density function can navigate almost all initial conditions from the initial set to the target set while avoiding the obstacle set. We show navigation results for simple integrator dynamics in complex environments as well as high-dimensional configuration spaces. Similarly, navigation results for obstacle avoidance involving robotics systems such as the two-link planar robotic arm manipulator are presented.

The rest of the paper is organized as follows. Section \ref{prelims} discuss the preliminaries and the problem formulation. Section \ref{sec:construction_nav_dens} discusses the construction of density functions, and Section \ref{sec:nav_dens} discusses the properties of density functions for the navigation problem. This is followed by application to robotic systems in section \ref{application_to_rob_sys} and conclusive remarks about the results in section \ref{conclusion}.

\section{Notations and Problem Statement} \label{prelims}

\noindent {\bf Notations}: The following notations will be used in this paper.  $\mathbb{R}^n$ denotes the $n$ dimensional Euclidean space, $\bx \in \mathbb{R}^n$ denotes a vector of system states, $\bu \in \mathbb{R}^n$ is a vector of control inputs. Let $\bX \subset \mR^n$ be a bounded subset that denotes the workspace for the robot. $\bX_0, \bX_T, \bX_{u_k} \subset \bX$,  for $k=1,\ldots, L$  denote the initial, target, and unsafe sets, respectively. With no loss of generality, we will assume that the target set is a single point set and located at the origin, i.e., $\bX_T=\{0\}$. $\bX_u=\cup_{k=1}^L\bX_{u_k}$ defines the unsafe set and $\bX_{s}:=\bX\setminus \bX_u$ defines the safe set. We will denote by $\bX_1:=\bX\setminus {\cal B}_{\delta}$, where ${\cal B}_\delta$ is the $\delta$ neighborhood of the origin for arbitrary small $\delta$.
 We use $\mathcal{C}^k(\bX)$ to denote the space of all $k$-times differentiable functions of $\bx$. We use ${\cal M}(\bX)$ to denote the space of all measures on $\bX$ and $m(\cdot)$ to denote the Lebesgue measure. $\mathds{1}_A(\bx)$ denotes the indicator function for set $A\subset \bX$.

The formal statement of the navigation problem that we solve in this paper is stated as follows. 
\begin{problem}(Almost everywhere navigation problem)\label{problem1} 
The objective of this problem is to design a smooth feedback control input $\bu=\bk(\bx)$ to drive the trajectories of the dynamical system
 \begin{align}\label{sys}
 \dot \bx=\bu,\;\;\;\;\;
 \end{align}
from almost every initial condition (w.r.t. Lebesgue measure)  from the initial set $\bX_0$ to the target set $\bX_T$ while avoiding the unsafe set $\bX_u$. 
\end{problem}
\begin{assumption} We  assume that there exists a feedback controller that solves the a.e. navigation problem as stated above.
\label{assum:feasibility}
\end{assumption}





\section{Construction of Density Function}\label{sec:construction_nav_dens}

The a.e. navigation problem, as stated in Problem \ref{problem1}, is solved using the navigation density function. The construction of the navigation density is inspired by the work of \cite{vaidya2018optimal, rantzer2001dual, vaidya2008lyapunov}. The navigation measure, as introduced in \cite{vaidya2018optimal}, has a physical interpretation of occupancy, where the measure of any set is equal to the occupancy of the system trajectories in the set, as shown in Figure \ref{fig:dens_nav_diagram}. Hence, zero occupancy in a set implies system trajectories not occupying that particular set. So by ensuring that the navigation measure is zero on the obstacle set and maximum on the target set, it is possible to induce dynamics whereby the system trajectories will reach the desired target set while avoiding the obstacle set. We exploit this occupancy-based interpretation in the construction of analytical density functions.

We start with the construction of the unsafe set, where the boundary of the unsafe set is described in terms of the zero-level set of a function.
Let $h_k(\bx)$ be a continuous scalar-valued function for $k=1,\ldots, L$ such that the  set $\{\bx\in \bX: h_k(\bx)\leq 0\}$, is connected with only one component.  Thus, the unsafe set $\bX_{u_k}$ is defined using the function $h_k(\bx)$ as follows
\begin{align}
\bX_{u_k}:=\{\bx\in \bX: h_k(\bx)\leq 0\}.
\end{align}


Next, we define a transition region $\bX_{s_k}$, which encloses the unsafe set $\bX_{u_k}$. Let $s_k(\bx)$ be a continuous scalar-valued function for $k=1,\ldots, L$ such that the set $\{\bx\in \bX: s_k(\bx)=0\}$ defines the boundary of this transition region. Then the transition region can be defined by the following set
\begin{align}
\bX_{s_k}:= \{\bx\in \bX : s_k(\bx)\leq 0\} \setminus \bX_{u_k}. \label{eq:transition_region}
\end{align}

\begin{figure}[tbp]
     \centering
     \includegraphics[width = 1\linewidth]{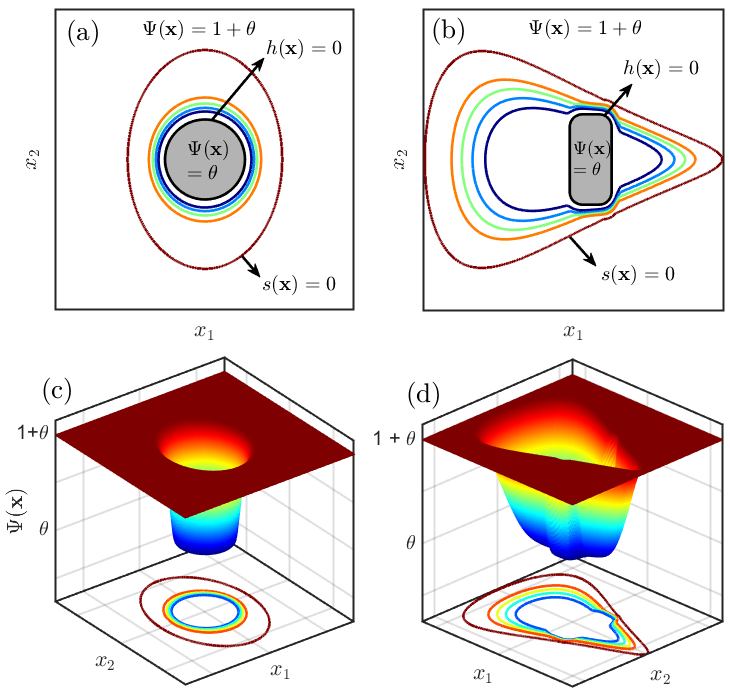}
     \caption{$\Psi(x)$ for (a) $\bX_u$ as a circle ($h(\bx) = ||\bx||^2 - r_1^2 \leq 0$) and transition region boundary as an ellipse ($s(\bx) = ||\mathbf{a}\bx||^2-r_2^2=0$ where $r_2 > r_1$ and $\mathbf{a}$ is a scaling vector), (b) $\bX_u$ as a rounded square ($h(\bx) = ||\bx||^4 - r_1^4 \leq 0$) and a transition region $(s(\bx) = a^2x_1^2 + b^2x_2^2c^{x_1} - r_2^2$ where $r_2 > r_1$; $a, b, c$ are parameters) defined using equation \eqref{density_fun}, (c-d) 3D view of (a) and (b) respectively.}
     \label{fig:psi_fig}
 \end{figure}

The proposed navigation density function is assumed to be of the form 
\begin{align}
\rho(\bx)=\frac{\prod_{k=1}^L \Psi_k(\bx)}{V(\bx)^\alpha}\label{density_fun}.
\end{align}
Here, the function $V(\bx)$ is the distance function that measures the distance from state $\bx$ to the target set, (i.e., the origin), and $\alpha$ is a positive scalar. In this paper, we assume $V(\bx)$ to be of the form $V(\bx)=\|\bx\|^2$. Additionally, $\Psi_k(\bx)$ is a smooth $\mathcal{C}^\infty$ function that captures the geometry of the unsafe set $\bX_{u_k}$ and can be constructed using the following sequence of functions. We first define an elementary $\mathcal{C}^\infty$ function $f$ as follows 
\begin{align} \label{eq:elementary_f}
    &f(\tau) = \begin{cases}
        \exp{(\frac{-1}{\tau})}, &\tau > 0 \\
        0, & \tau \leq 0
    \end{cases},
\end{align}
where $\tau \in \mathbb{R}$ \cite{tu2011manifolds}. Next, we construct a smooth version of a step function $\bar{f}$ from $f$ as follows
\begin{equation} \label{eqn:smooth_step_g}
    \bar{f}(\tau) = \frac{f(\tau)}{f(\tau)+f(1-\tau)}.
\end{equation}
Here, $\bar{f}$ serves as the elementary function for representing zero and nonzero occupation through density. Furthermore, the form of the elementary function, $\bar f$, is chosen to ensure that the gradient of the density function is well-defined. To incorporate more general geometric information about the environment, we define a change of variables such that $\phi_k(\bx) = \bar{f}\Bigl(\frac{h_k(\bx)}{h_k(\bx) - s_k(\bx)}\Bigr)$. The resulting function $\Phi_k(\bx)$ take the following form,
\begin{align} \label{eq:inverse_bump}
\Phi_k(\bx) = \begin{cases}
    0, & \bx \in \bX_{u_k}  \\
    \phi_k(\bx), & \bx \in \bX_{s_k}\\
    1, & \rm{otherwise}.
\end{cases}
\end{align}
Finally, the function $\Psi_k(\bx)$ is defined as
\begin{align}
\Psi_k(\bx)=\Phi_k(\bx)+\theta,\label{inverse_bump}
\end{align}
where $\theta>0$ is some positive parameter. The parameters $\theta$ and $\alpha$ are introduced in the construction of the navigation density. The physical significance of these parameters and the assumption made on these parameters and functions are stated in the following remark.
\begin{remark}\label{remark_psi} 
\begin{itemize}
\item The distance function $V(\bx)$ can be modified to adapt to the geometry of the underlying configuration space. For a Euclidean space with $\bx \in \mathbb{R}^n$, we pick $V(\bx) =\|\bx\|^2$. 

\item The parameter $\alpha$ is used to control the sharpness of the distance function and is used in the proof of the main convergence results.

\item The function $\Psi_k(\bx)$ is a $\theta$ shifted version of inverse bump function $\Phi_k(\bx)$ and hence strictly positive i.e., $\Psi_k(\bx)\geq \theta>0$ for $k=1,\ldots, L$. 

\item $\Psi_k(\bx)$ makes a smooth transition from $\theta$ to $1+\theta$ in the transition region $\bX_{s_k}$. 

\item The transition region, $\bX_{s_k}$, acts as a sensing region for system trajectories where they start to react to the unsafe set. We refer to the transition region as the sensing region for the rest of this paper.

\item $h_k(\bx)=0$ defines the boundary of the unsafe set and $s_k(\bx)=0$ defines the boundary of the sensing region. Refer to Figure \ref{fig:psi_fig} for an illustrative example. In the simplest case, the function $s_k(\bx)$ can be chosen to be synonymous to $h_k(\bx)$, such that $h_k(\bx) - s_k(\bx) = \sigma $ (where $\sigma > 0$ is a constant) uniformly scales the unsafe set to form a sensing region.



\end{itemize}
 
\end{remark}


We assume explicit bounds on the functions $\Psi_k$, $V$, and their derivatives which follow from the construction of the density function in equation \eqref{density_fun}. It is important to emphasize that it is not necessary to estimate these bounds, but the existence of these bounds is used as part of the proof of the main results of this paper.
 \begin{assumption}\label{assume_main} 
 \begin{enumerate}

\item[] 
\item We assume that the distance between the initial set, the target set, and the unsafe sets are all bounded away from zero by some positive constant, say $\zeta$.

\item For $\bx\in \bX_{u_k}$, let
\begin{align}V_{min}^k=\min_{\bx\in \bX_{u_k}}V(\bx)>0.\;\;\;
\label{eq:V_min}
\end{align}
Since the distance between the unsafe set and the target set is bounded away from zero, the above quantity is well-defined and greater than zero.
\item Furthermore, $m(\bX_{u_k})$, i.e., the Lebesgue measure of the unsafe set, is assumed to be finite, with $\theta$ satisfying the following inequality for any given $\epsilon>0$,
\begin{align}\theta\leq \frac{V_{min}^k}{m(\bX_{u_k})} \epsilon ,\;\;\;\;k=1,\ldots, L.\label{assume_epsilon}
\end{align}
\[ \underline{c}_V \leq V(\bx)\leq \bar c_V, \;\;\; \underline{c}_{V_x}\leq \left|\frac{\partial V}{\partial x_j}\right| \leq \bar c_{V_x}, \;\;\; {\color{black} \left|\frac{\partial^2 V}{\partial x_j^2} \right|\leq \bar c_{V_{x^2}}}\]
\[\left| \frac{\partial \Psi}{\partial x_j}\right|\leq \bar c_{\Psi_x}, \;\;\;\; \left| \frac{\partial^2 \Psi}{\partial x_j^2}\right|\leq \bar c_{\Psi_x^2},\;\;\;\;{\color{black}j=1,\ldots,n}.\]
Further, by construction, both the first and second derivatives of $\Psi$ w.r.t. $x_j$ are zero outside the transition region

\item Outside the transition region and in 
$\bX_1$, we assume 
\[\frac{\partial^2 V}{\partial x_j^2}\leq \bar d_{V_x^2},\;\;V\leq \bar{d}_V \|\bx\|^2,\;\;\left|\frac{\partial V}{\partial x_j}\right|\geq \underline{d}_{V_x}\|\bx\|\;j=1,\ldots,n.\]
We have used lower bar, $\underline{c},\; \underline{d}$, and upper bar, $\bar c,\; \bar d$, notations to help define the lower and upper positive bounds on functions. The subscripts for $c$ and $d$ signify the corresponding functions.

\end{enumerate}

\end{assumption}

\section{Almost Everywhere Navigation Using Density Functions} \label{sec:nav_dens}

Given the construction of $\rho(\mathbf{x})$ in (\ref{density_fun}), we design a controller for navigation as the positive gradient of the density function $\rho(\bx)$, i.e.,
\begin{align}
&\dot \bx=\bk(\bx) = \nabla \rho(\mathbf{x}) \nonumber \\ 
&=\Biggl(-\frac{\alpha}{V^{\alpha+1}}\frac{\partial V}{\partial \bx}\prod_{k=1}^L \Psi_k(\bx)
+\frac{1}{V^\alpha}\frac{\partial }{\partial \bx}\prod_{k=1}^L \Psi_k(\bx)\Biggr)^\top.
\label{system_navigation}
\end{align}

\begin{remark} \label{remark:local_controller}
 {\color{black} We make following modification to 
(\ref{system_navigation}) to ensure that the vector field is well-defined and the origin is locally asymptotically stable in ${\cal B}_\delta$.    $\dot \bx = \left[1-\Bar{f}(\tau)\right]\nabla \rho(\bx) - \Bar{f}(\tau)\bx$ where, $\bar f$ is as defined in (\ref{eqn:smooth_step_g}).} {\color{black} With this modification, we will continue to work with (\ref{system_navigation}) with the assumption that the origin is locally asymptotically stable in ${\cal B}_\delta$ for (\ref{system_navigation}). }
\end{remark}





The main result of the paper is given in the following theorem.
\begin{theorem} \label{theorem_main}Under Assumptions \ref{assum:feasibility} and \ref{assume_main}, the dynamical system (\ref{system_navigation}) will solve the a.e. navigation problem as  stated in Problem \ref{problem1}. 
\end{theorem}

Proof of this main theorem is differed to the Appendix. 
The feedback controller design for the a.e. navigation problem is illustrated in pseudo-code in Algorithm \ref{alg:density_based_nav}.

\begin{algorithm}
\caption{Density-based Navigation Algorithm}\label{alg:density_based_nav}
\begin{algorithmic}
\INPUT{$\mathbf{X_0, X_u, X_T}$}
\item[\quad $\Psi(\mathbf{x}) \leftarrow 1$]
\item[\quad Define $V(\bx)$ according to configuration]
\For{$X_{u_k}$ in $\mathbf{X_u}$}
    \DEFINE $h_k(\mathbf{x})$ and $s_k(\mathbf{x})$ (see Remark \ref{remark_psi} and \ref{remark_tuning})
    \item[\qquad Form $\Psi_k(\mathbf{x})$ from $h_k(\mathbf{x})$ and $s_k(\mathbf{x})$ (see equation \ref{eq:inverse_bump})]
    \item[\qquad $\Psi(\mathbf{x})$ $\leftarrow$ $\Psi(\mathbf{x}) \times \Psi_k(\mathbf{x})$]
\EndFor

\item[\quad $\rho(\mathbf{x}) = \frac{\Psi(\mathbf{x})}{V(\mathbf{x})^\alpha}$]
\item[\quad $u = \nabla \rho
(\mathbf{x})$]

\end{algorithmic}
\end{algorithm}
The rest of the section showcases the navigation results using the controller designed from the analytical density function. We first show the characteristics of the proposed controller, which validates the a.e. navigation properties. Then, we extend our feedback controller to a more complex environment. Lastly, a comparison of our algorithm to NFs is presented. 

\subsection{Characteristics of Density Functions}
In this example, we demonstrate the a.e. navigation properties of the proposed controller. The navigation problem is defined with the target set at $\bX_T = (4,-3)$ and the unsafe set $\bX_u$, which is constructed using a circular inverse bump function with {\small$h(\bx) = ||\bx||^2 - r_1^2$} and {\small$s(\bx) = ||\bx||^2 - r_2^2$} with $r_1 = 2$ and  $r_2 = 3$. Hence, $\bX_{s_k}$ for the inverse bump function is defined on the domain $2< \|\bx\|<3$.

\begin{figure}
    \centering
    \includegraphics[width = \linewidth]{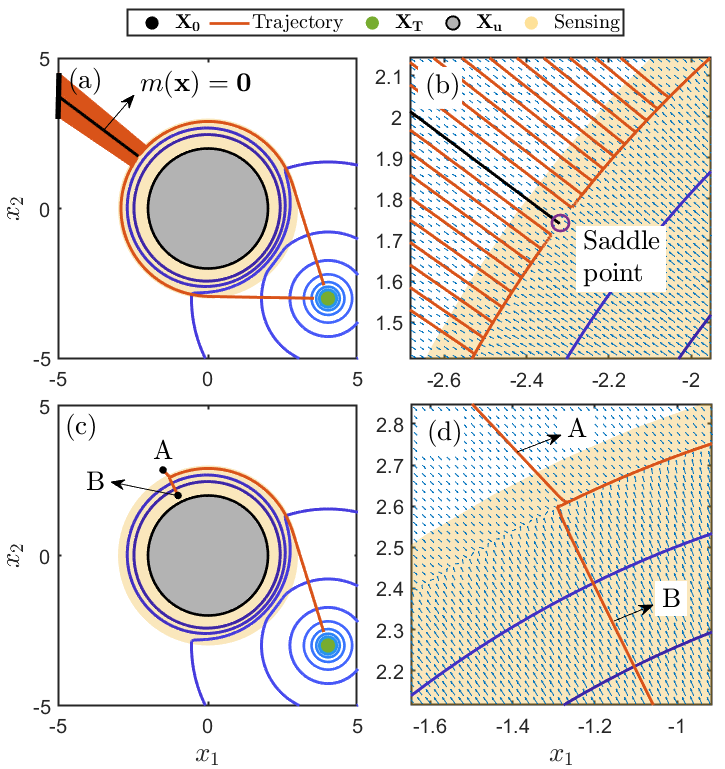}
    \caption{(a) Trajectories converge to the target set (green) while avoiding the unsafe set (gray) with a.e. convergence, (b) Initial conditions along the zero-measure set (black) converge to a saddle point (purple), (c) Trajectories starting at A ($s(\bx)>0$) and B (in $\bX_{s_k}$) converge to the target set, (d) Trajectories starting from A and B follow the same path near the boundary of $s(\bx)$.}
    \label{fig:saddle_points}
\end{figure}

Figure \ref{fig:saddle_points}a illustrates the a.e convergence of the proposed controller with initial conditions set defined by a line at the top left of the environment boundary. The blue contour lines represent the level sets of the density function. For this example, all the initial conditions starting on the set $\{\bX_0 \subset \bX: m(\bX_0) = 0\}$, which is polar opposite of the target set, cannot converge. This set of initial conditions constitutes a measure zero set. Furthermore, these initial conditions are attracted to a saddle point, implying the existence of local maxima (shown in Figure \ref{fig:saddle_points}b). Note that the existence of a saddle point will imply the existence of local maxima. Any other trajectory starting from an initial condition perturbed from the zero-measure set converges to the target set $\bX_T$ while avoiding the obstacle set $\bX_{u}$. Furthermore, we look at the characteristics of initial conditions starting outside the sensing region, defined as a state $\bx$ such that $s(\bx) \geq 0$ (trajectory A), and within the sensing region, defined as a state $\bx$ such that $0 < h(\bx) < s(\bx)$ (trajectory B),  shown in Figure \ref{fig:saddle_points}c. The gradients of the density function $\rho(\bx)$ are such that trajectory A starts to react as it enters the sensing region while trajectory B is repelled outward towards the boundary of the sensing region before converging to the target set (see Figure \ref{fig:saddle_points}d).

\subsection{Complex environment}
\begin{figure}
     \centering
     \begin{subfigure}[b]{\linewidth}
         \centering
         \includegraphics[width=\linewidth]{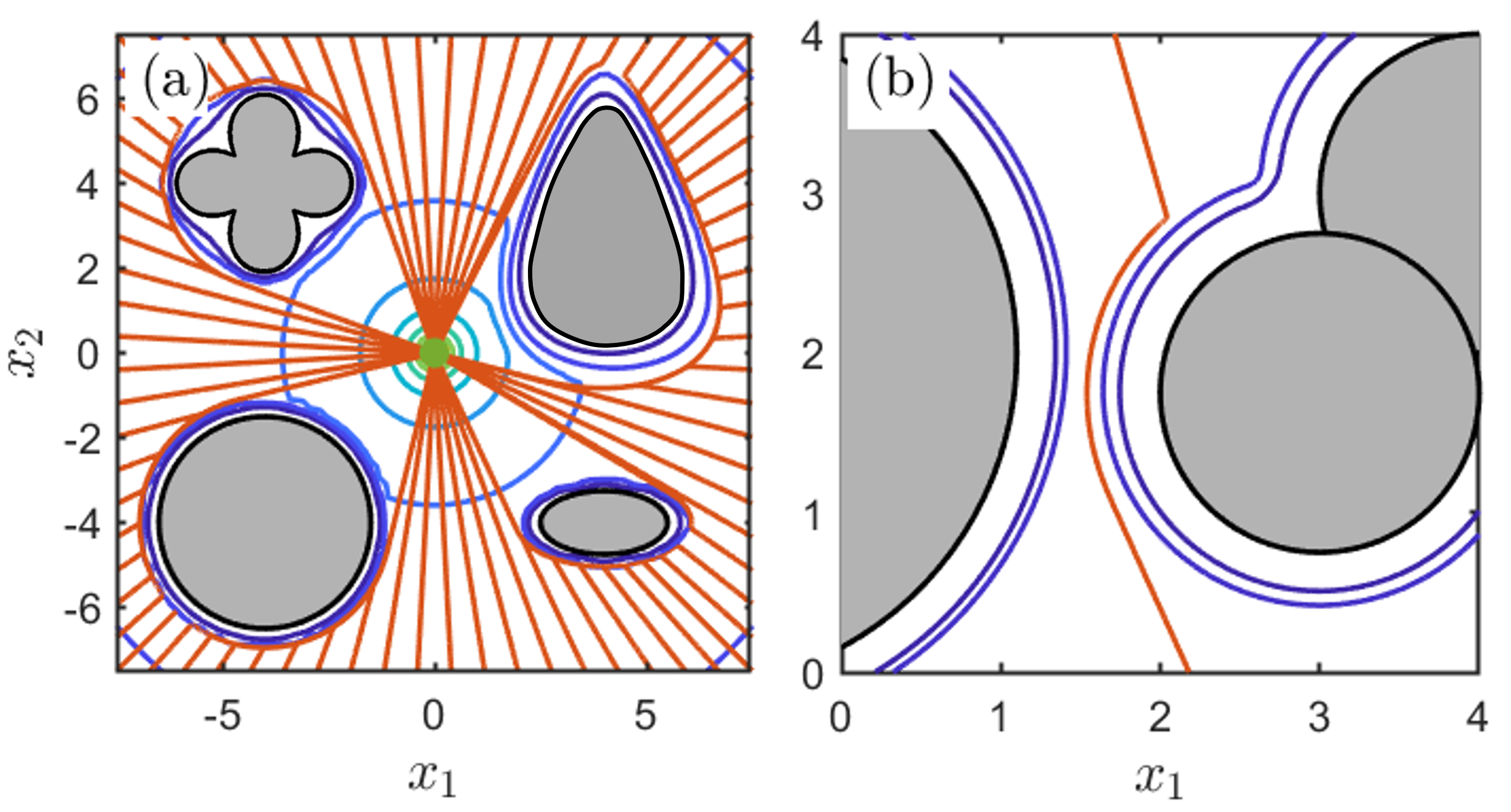}
     \end{subfigure}
     \hfill
     \begin{subfigure}[b]{\linewidth}
         \centering
         \includegraphics[width=\linewidth]{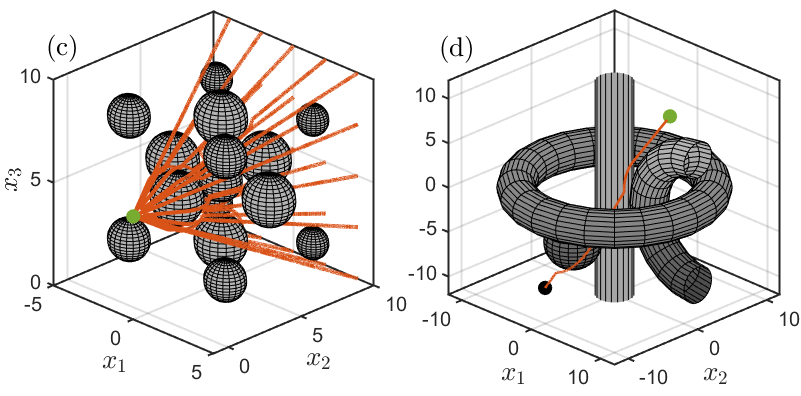}
     \end{subfigure}
    \caption{(a) Trajectories converge to the target set (green) while avoiding arbitrary obstacles (gray), (b) Trajectory finding a narrow feasible region around obstacles, (c) Navigation in a spherical grid, (d) Navigation through two tori, unbounded cylinder, and sphere.}
    \label{fig:complex_env}
\end{figure}

One of the main features of our proposed navigation density is that it can incorporate complex shapes of the obstacle set, which is captured in terms of the unsafe set by some appropriate function 
$h_k(\bx)$.
The unsafe set {\small$\bX_u \in \mathbb{R}^2$} in Figure \ref{fig:complex_env}a is constructed using an implicit function that geometrically represents a circle, an ellipse, an oval, and a bowtie. We show that the initial conditions starting along the boundary converge to the goal at the center while safely avoiding obstacles. The proposed controller can also satisfy a.e navigation in complex maze-like environments. Figure \ref{fig:complex_env}b shows a trajectory finding a tight feasible region between two obstacles while navigating to the target set. Furthermore, this can be easily extended to navigation problems in higher dimensions. Figure \ref{fig:complex_env}c shows all trajectories starting from a plane converging to the target set while avoiding obstacles represented as 3D spheres. Figure \ref{fig:complex_env}d shows navigation with unsafe sets composed of two tori, an unbounded cylinder, and a sphere. We note that unlike \cite{ filippidis2012navigation, loizou2011closed}, the construction of the density function naturally admits any complex shapes.

\subsection{Comparison to Navigation Functions}\label{comparison}
\begin{figure}
    \centering
    \includegraphics[width = \linewidth]{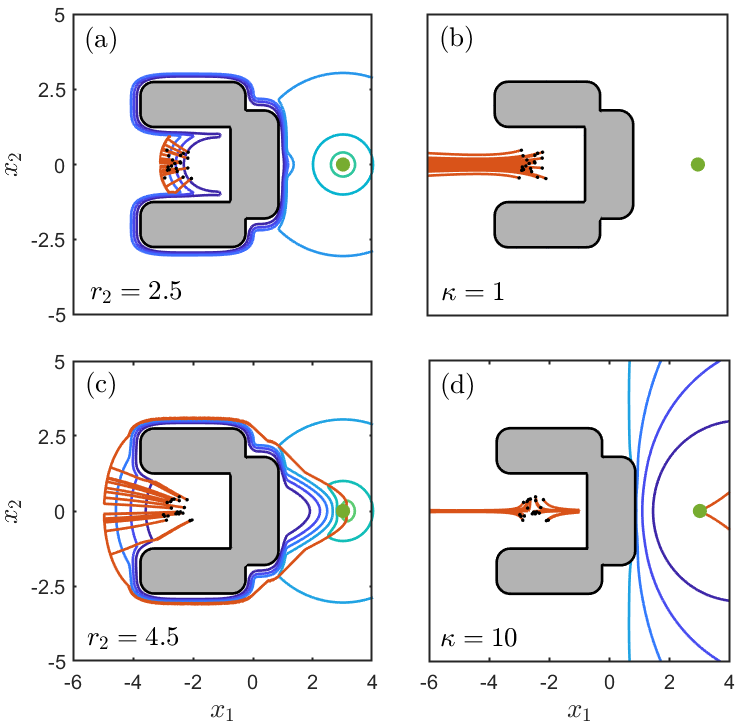}
    \caption{Comparison of density functions and NFs for random initial conditions. The sensing region for the density function is defined by $s(\bx) = a^2x_1^2 + b^2x_2^2c^{x_1} - r^2$ ($r$, $a, b, c$ are parameters). For (a) $r = 2.5$, trajectories don't converge, while setting (c) $r = 4.5$ leads to all trajectories converging. NFs with their corresponding tuning parameter for convergence (b) $\kappa=1$ and (d) $\kappa=10$ lead to trajectories not converging.}
    \label{fig:densityVsNf}
\end{figure}

In this section, we compare the a.e. convergence property of artificial potential field NF to the proposed density functions in a complex environment as shown in Figure \ref{fig:densityVsNf}. More specifically, we compare the tuning of $s(\bx)$ for a.e. convergence in the density function formulation shown in equation \eqref{density_fun} to the tuning of $\kappa \in \mathbb{R}$ for a.e. convergence in NFs proposed in \cite[Ch.~3, p.~36]{rimon1990exact},
\begin{equation}
    \psi_k(\bx) = \frac{||\bx-\bx_g||^2}{||\bx-\bx_g||^2 + \beta({\bx})^{1/\kappa}},
\end{equation}
where $\bx_g$ is the desired goal location, {\small$\beta({\bx})$} is an obstacle function and $\kappa$ is a tuning parameter. 

Although a domain is not necessary in the density formulation, NFs do require a radially bounded sphere world. Hence, we define an appropriate bounded sphere world of radius 25. The authors note that NFs do not make any claims about tuning $\kappa$ for a.e. convergence other than the sphere world and its extensions \cite{rimon1990exact, filippidis2012navigation}, but for the sake of comparison, we look at an environment with a C-shaped unsafe set. We then look at initial conditions that lie inside the C-shaped unsafe set with the target set defined outside the cavity of the unsafe set.

Figures \ref{fig:densityVsNf}a and \ref{fig:densityVsNf}b show that the trajectories do not converge to the goal for all random initial conditions for small values in tuning parameter in either the density function formulation or the artificial potential field NF formulation. This is expected in NF as only large $\kappa$ in a sphere world guarantees a.e. convergence. Likewise, the density formulation sees the same results. However, tuning $s(\bx)$ such that the density function formulation has a.e. convergence property is intuitive, as stated below in Remark \ref{remark_tuning}. This is shown in Figure \ref{fig:densityVsNf}c, where tuning $s(\bx)$ to be larger than the C-shaped unsafe sets results in all system trajectories converging to the target set. Note, no explicit mapping to a simplistic unsafe set (e.g., circle) is required, where the same cannot be stated for NFs (even with high $\kappa$), which does not give a.e. convergence results for complex unsafe sets. This can be seen in Figure \ref{fig:densityVsNf}d, where some trajectories exit the unsafe set and converge to the goal (by taking a large curvature path) while others get trapped inside the cavity of the unsafe set.

\begin{remark} \label{remark_tuning} The tuning parameter in the design of the navigation density functions are $\alpha$, and $s_k(\bx)$. The tuning of $\alpha$ depends on the rate of convergence of the trajectories. Although a large value of $\alpha$ is required for a.e. navigation (as shown in Appendix), in practice, even small values of $\alpha$ (between 1 to 10) have shown to work. The tuning of $s_k(\bx)$ is physically intuitive, as it signifies the sensing region. Hence, a sensing region that encompasses the unsafe set with a sufficiently curved convex set has worked in the simulations.
 \end{remark}

\section{Application to Robotic Systems}\label{application_to_rob_sys}
We consider cases which are highly important in application, with specifics to robotic systems. These cases, constrained control, stochastic settings, and fully actuated multi-body systems are considered in the subsequent sections.

\subsection{Constrained Control w/ Density Function}
The case of controlling the magnitude of the controller defined in \eqref{system_navigation} is highly crucial in practical systems due to actuation limits. Although, the magnitude of the controller can be implicitly controlled through the tuning of $\alpha$, as change in $\alpha$ changes the sharpness of $V(\bx)$, hence change in gradient  of density (i.e. change in magnitude of control), we consider explicitly defining control constraints. In particular, we consider a system with constraints in the following form
\begin{equation}
    \dot \bx = \bu = \nabla \rho(\bx), \quad \bu \in [-u_{max}, u_{max}],
    \label{eq:constrained_nav_dens}
\end{equation}
where $u_{max}$ is the bound on control. Without formality, we constrain the control when $||\bu||_{\infty} > u_{max}$ by normalizing the control
\begin{equation}
    \bar \bu = \frac{\bu}{||\bu||_\infty}u_{max},
\end{equation} 
where $\bar \bu$ is the constrained control. 

\begin{figure}[htbp]
    \centering
    \includegraphics[width = \linewidth]{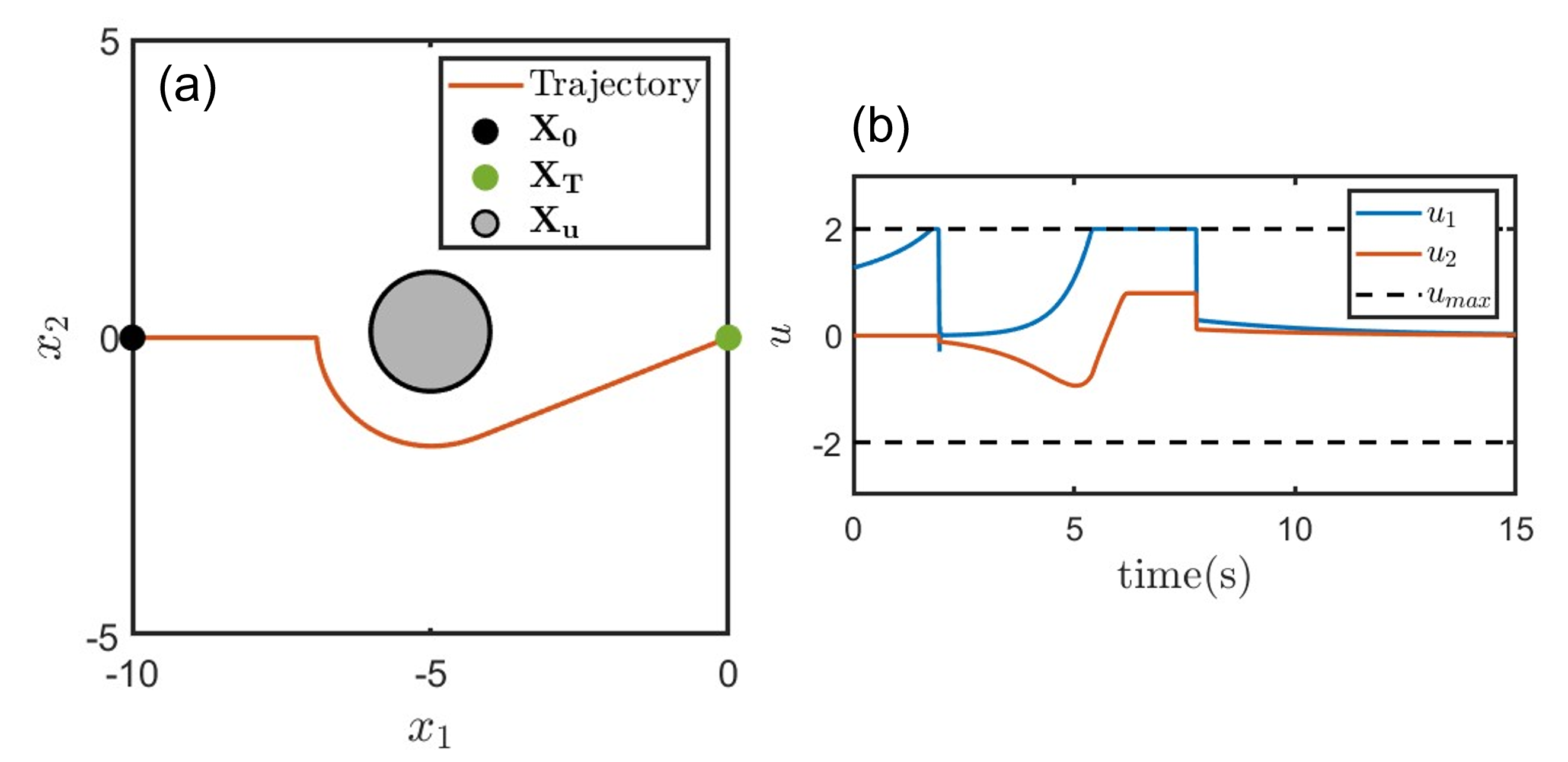}
    \caption{Constrained control w/ navigation density}
    \label{fig:constrained_density}
\end{figure}

\subsection{Performance of Density Function w/ Noise}
\begin{figure}[htbp]
    \centering
    \includegraphics[width = \linewidth]{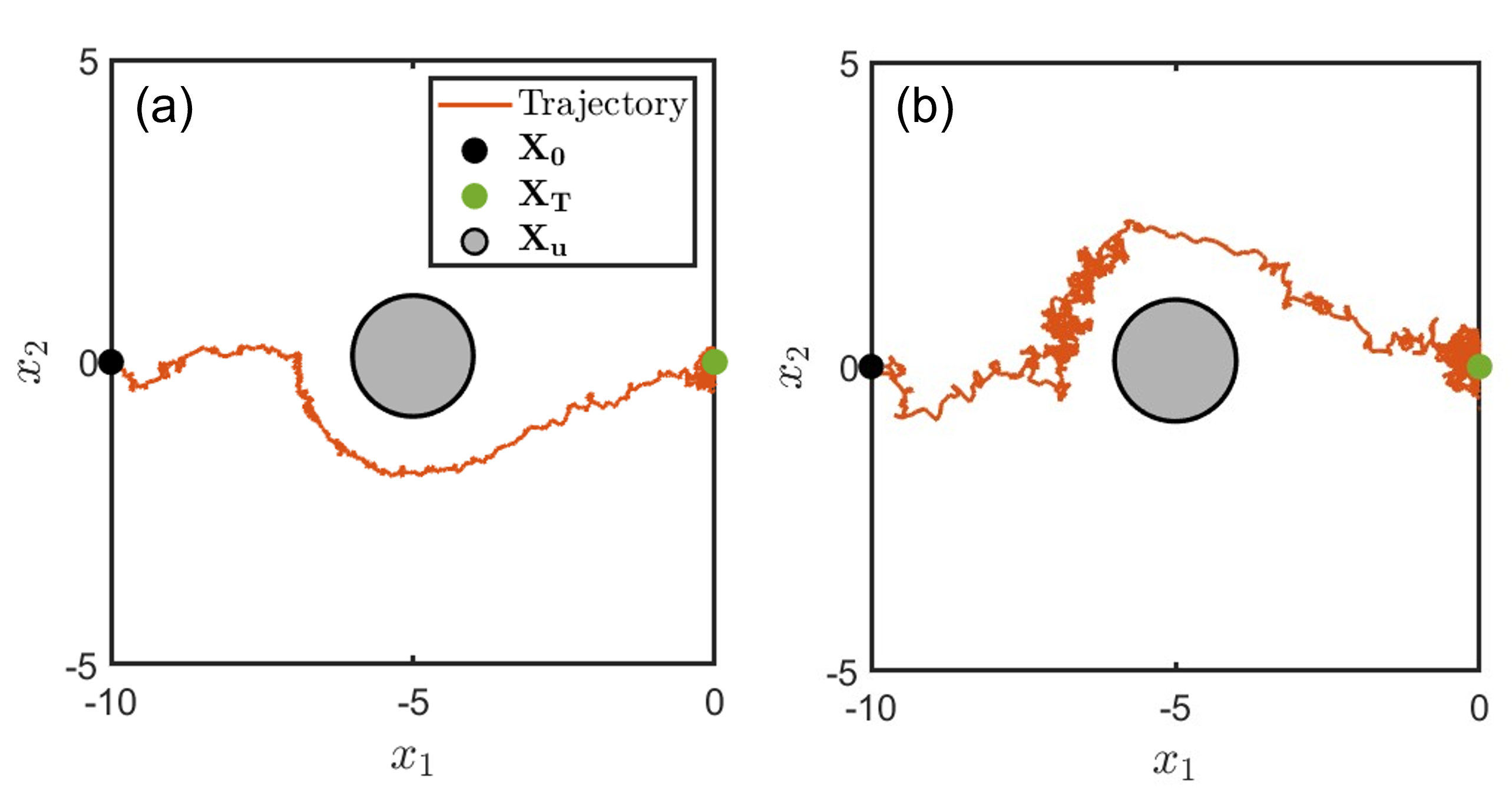}
    \caption{Density function w/ gaussian white noise of $\mu = 0$ and (a) $\Sigma = 10^{-3} \times I_2$ (b) $\Sigma = 5 \times 10^{-3} \times I_2$}
    \label{fig:density_noise}
\end{figure}

We consider the performance of our controller in a stochastic setting where noise is entered through the control input
\begin{equation}
    \dot \bx = \bu + \bw \quad \bu \in [-u_{max}, u_{max}],
\end{equation}
where $\bw \in \mathcal{N}(\mu, \Sigma)$ is the gaussian white noise with mean $\mu = 0$ and covariance $\Sigma$. Figure \ref{fig:density_noise} showcases the navigation problem with control noise for varying levels of covariance.

We see that the feedback controller is capable of invariance while converging towards the goal with noise. Although the invariance of our control law is not guaranteed, we see that up to a certain bound on the noise, the control performance is robust.

\subsection{Fully Actuated Robotic System}
We also extend the density function presented in Section \ref{sec:construction_nav_dens} to a general class of fully actuated robotic systems. For a robot with $n$ joints and $n$ rigid links, the system's dynamics can be expressed using the Euler-Lagrange equations. Consider an unconstrained system where $\bf{M}(\bf{q})$ is the inertia matrix and $\bf{H}(\bf{q},\dot{\bf{q}})$ represents the Coriolis and gravity effects on the system, $\bf{q}$ $\in {\mathbb S}^1 \times {\mathbb S}^1$. Then the corresponding system is represented as follows
\begin{equation}
   \bf{M}(\bf{q})\ddot{\bf{q}} +\bf{H}(\bf{q},\dot{\bf{q}}) = \bf{u}.
   \label{eq:eulerLagragnge}
\end{equation}
We then take a similar approach outlined in \cite{KoditschekMotion} in which there exists an equivalent "planning" system defined by $\dot{\bf{q}} = \nabla \rho(\mathbf{q})$ and a control law given by $\bu = \nabla \rho(\mathbf{q}) + d(\bf{q},\dot{\bf{q}})$
($d(\bf{q},\dot{\bf{q}})$ is a dissipative term and $\dot{\bf{q}}^\top d(\bf{q},\dot{\bf{q}}) < 0$), where the system defined in \eqref{eq:eulerLagragnge} tracks the planning system asymptotically \cite{KoditschekMotion}. For a general robotic system such as the system defined in equation \eqref{eq:eulerLagragnge}, $d(\bf{q},\dot{\bf{q}})$ can be selected such that it cancels out the nonlinearities of the system similar to the inverse dynamics approach. Therefore, we define a density-based inverse dynamics controller given by
\begin{equation}
    \bf{u}_\rho = \bf{M}(\bf{q})\ddot{\bf{q}}_d + \bf{H}(\bf{q},\dot{\bf{q}}) + \bf{M}(\bf{q}) \Biggl(\bf{K}_p \nabla \rho(\mathbf{e})  - \bf{K}_v \bf{\dot{e}} \Biggr),
\end{equation}
where $\be := \bf{q}-\bf{q}_d$, $\dot{\be} := \bf{\dot{q}}-\bf{\dot{q}}_d$, $\bf{q}_d$ is the desired reference trajectory to follow, and $\bf{K}_p$ and $\bf{K}_v$ are positive definite gain matrices.
\begin{figure}
    \centering
    \includegraphics[width = \linewidth]{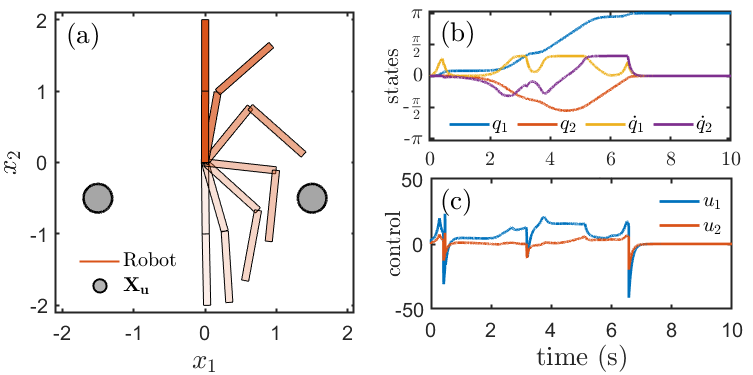}
    \caption{(a) Robot (red) converges to the goal ($\pi,0$) starting from equilibrium (0,0) while avoiding obstacles (gray). (b) state trajectories of the robot and (c) control inputs for executing the swing-up maneuver.}
    \label{fig:swingup}
\end{figure}
Figure \ref{fig:swingup}a shows a fully actuated two-link planar robotic arm executing a swing-up maneuver with $\bf{K}_p$ = ${\rm diag}([1,1]),\;\;\;$ $\bf{K}_v$ = ${\rm diag}([10,10])$ and $V({\bf q}) = (1-\cos(q_1)(1-\cos(q_2))$. The mass and length of each link are set to unity. The task space obstacles (circular with a radius of 0.2) are mapped to joint space and approximated using inverse bump functions. The reference trajectories are obtained in joint space based on the planning system $\dot{\bf{q}} = \nabla \rho(\mathbf{q})$. The corresponding state and control trajectories are shown in Figures \ref{fig:swingup}b and \ref{fig:swingup}c, respectively. It is seen that the density-based inverse dynamics controller drives the two-link manipulator to the upright position while avoiding the obstacle set.


\section{Conclusions}\label{conclusion}
This work provides an analytical construction for the navigation density. Moreso, we prove that the navigation density solves the almost everywhere navigation problem. The proposed navigation density can be viewed as dual to the popular navigation function and is derived based on the occupancy-based interpretation of the density function. The navigation density has a few advantages compared to navigation functions. Unlike navigation functions, which are hard to construct, navigation density can be easily constructed. Furthermore, the density function formulation can incorporate arbitrary shapes of the unsafe set. We provide simulation results for navigation using density function in complex and high dimensional environments as demonstrated. Lastly, we also demonstrate the application of the density function for control on a robotic system with safety constraints.  
\section{Appendix}
The proof of Theorem \ref{theorem_main} relies on the following Lemma.
\begin{lemma}\label{lemma1} Consider the navigation density function as given in equation  (\ref{density_fun}), then under Assumption \ref{assume_main}, we have   
\begin{align}
&\nabla\cdot(\bk(\bx)\rho(\bx))\geq 0,\;\;\;a.e. \;\; \bx\in \bX, \label{eq:rantzer_cond}\\
&\nabla\cdot (\bk(\bx)\rho(\bx))\geq \xi>0 \;\;{\rm for}\;\;\bx\in \bX_0,\label{honX0}
\end{align}
where $\bk(\bx)=\nabla \rho(\bx)$ is the feedback control input as given in equation (\ref{system_navigation}). 
\end{lemma}
{\bf Proof: } We have 
\begin{align}
\nabla\cdot(\bk(\bx)\rho(\bx))=\rho(\bx)\nabla\cdot \bk(\bx)+\frac{\partial \rho}{\partial \bx}\frac{\partial \rho}{\partial \bx}^\top.
\end{align}
Since $\rho(\bx)>0$ and $\frac{\partial \rho}{\partial \bx}\frac{\partial \rho}{\partial \bx}^\top\geq 0$, the proof will follow  if we can show that  $\nabla\cdot \bk(\bx)\geq 0$. We have 
\begin{align}
\nabla\cdot \bk(\bx)=\sum_{j=1}^n \frac{\partial^2\rho}{\partial x_j^2}.
\end{align}
Letting $\Psi(\bx)=\prod_{k=1}^L \Psi_k(\bx)$, we obtain
\begin{align}
\frac{\partial^2 \rho}{\partial x_j^2}&=\frac{\partial }{\partial x_j}\left(-\frac{\alpha}{V^{\alpha+1}}\frac{\partial V}{\partial x_j}  \Psi(\bx)+\frac{1}{V^\alpha}\frac{\partial \Psi}{\partial x_j}\right) \nonumber \\
&=\frac{\alpha}{V^\alpha}\left(\frac{(\alpha+1)}{V^{2}}\left|\frac{\partial V}{\partial x_j}\right|^2{\color{black}-\frac{1}{V}\frac{\partial^2 V}{\partial x_j^2}}\right)\Psi(\bx) \nonumber \\
&+\frac{\alpha}{V^\alpha}\left(-\frac{2}{V}\frac{\partial V}{\partial x_j}\frac{\partial \Psi}{\partial x_j}{\color{black}+}\frac{1}{\alpha}\frac{\partial \Psi^2}{\partial x_j^2}\right).\label{ee2}
\end{align}
It is important to note that the last two terms in the above expression are non-zero only in the transition region $\bX_{s_k}$. Outside this transition region $\frac{\partial \Psi}{\partial x_j}=0$ and $\frac{\partial^2\Psi}{\partial x_j^2}=0$. To show that the above quantity is positive outside the transition region in $\bX_1$, we use the bounds from Assumption \ref{assume_main}. We have
\[\frac{(\alpha+1)}{V}\left|\frac{\partial V}{\partial x_j}\right|^2-\frac{\partial^2 V}{\partial x_j^2}\geq (\alpha+1)\bar d^{-1}_V \underline{d}^2_{V_x}-\bar d_{V_x^2}\]
Thus, by choosing $\alpha$ sufficiently large, the above quantity can be made positive.

We next show that equation (\ref{ee2}) is non-negative in the transition region. For this, we make use of the following facts. First,  $\Psi_k(\bx)\geq \theta>0$ for $k=1,\ldots, L$ and hence $\Psi(\bx)$ is bounded away from zero. Second, from the construction of $\Psi(\bx)$ and $V(\bx)$ functions there exists uniform bounds on $\frac{\partial \Psi_k}{\partial x_j},\frac{\partial^2 \Psi_k}{\partial x_j^2}$, $\frac{\partial V}{\partial x_j}$ and $\frac{\partial^2 V}{\partial x^2_j}$. Third, using Assumption \ref{assume_main}, we know that the distance between the unsafe set and the target set is bounded away from zero by a positive constant $\zeta$ and hence $\left|\frac{\partial V}{\partial x_j}\right|^2$ is bounded away from zero. Hence, the following bounds can be obtained for the $\frac{\partial^2\rho}{\partial x_j^2}$ term
\begin{align}
&\frac{(\alpha+1)}{V^{2}}\left|\frac{\partial V}{\partial x_j}\right|^2 \Psi(\bx) \geq \left((\alpha+1) \bar c_{V}^{-2}\underline{c}_{V_x}^2 \right)\theta,\;\;\;\frac{1}{\alpha}\frac{\partial^2\Psi}{\partial x_j^2}\geq -\frac{\bar c_{\Psi_x^2}}{\alpha}, \nonumber \\
&{\color{black}-\frac{1}{V}\frac{\partial^2 V}{\partial x_j^2}}\Psi(\bx) \geq {\color{black}-\underline c_{V}^{-1}\bar{c}_{V_{x^2}}\theta},\;\;\; -\frac{2}{V}\frac{\partial V}{\partial x_j}\frac{\partial \Psi}{\partial x_j}\geq -2 \underline c_{V}^{-1} \bar c_{V_x}\bar c_{\Psi_x}.\nonumber
\end{align}
Therefore, we have following lower bound for $\frac{\partial^2 \rho}{\partial x_j^2}$
\begin{align}
\frac{\partial^2 \rho}{\partial x_j^2}\geq&\frac{\alpha}{V^\alpha}\left(\left((\alpha+1) \bar c_{V}^{-2}\underline{c}_{V_x}^2 {\color{black}-\underline c_{V}^{-1}\bar{c}_{V_{x^2}}} \right)\theta\right) \nonumber \\
&+\frac{\alpha}{V^\alpha}\left(-2  \underline{c}_{V}^{-1} \bar c_{V_x}\bar c_{\Psi_x}-\frac{\bar c_{\Psi_x^2}}{\alpha}\right).\nonumber
\end{align}
Hence, by choosing $\alpha$ sufficiently large, of order $\frac{1}{\theta}$, we can make the term inside the bracket positive.


To show that equation (\ref{honX0}) is satisfied, we again make use of Assumption \ref{assume_main} and the fact that $\Psi(\bx)=1$, $\frac{\partial \Psi}{\partial \bx_j}=0$, and $\frac{\partial^2 \Psi}{\partial x_j^2}=0$ for $\bx\in \bX_0$ and $j=1,\ldots, n$. Further, $\left|\frac{\partial V}{\partial x_j}\right|^2$ is bounded away from zero. Hence for,  $\bx\in \bX_0$, we obtain
\begin{align}
\nabla\cdot({\color{black}\bk(\bx)}\rho(\bx))=&\frac{\partial \rho}{\partial \bx}\frac{\partial \rho}{\partial \bx}^\top +\frac{\alpha(\alpha+1)}{V^{\alpha+2}}\left|\frac{\partial V}{\partial x_j}\right|^2 \nonumber \\
&{\color{red}-\frac{1}{V^{\alpha+1}}\frac{\partial^2 V}{\partial x_j^2}} \geq \xi>0. \nonumber
\end{align}
for some $\xi>0$.

{\bf Proof of Theorem \ref{theorem_main}: } Using the results of Lemma \ref{lemma1}, we know that the density $\rho$ satisfies 
\begin{align}\nabla\cdot (\bk(\bx)\rho(\bx))=g(\bx)\label{pde}
\end{align}
for some $g(\bx)\geq 0$ such that $g(\bx)\geq \xi>0$ for $\bx\in \bX_0$. 





Since $\rho(\bx)$ satisfies the linear partial differential equation (\ref{pde}), it follows using the method of characteristics that the solution $\rho(x)$ can be written in terms of the solution $\bs_t(\bx)$, of the system $\dot \bx=\bk(\bx)$ as follows \cite{rajeev_continuous_journal}
\begin{align}\rho(\bx)=\frac{\Psi(\bx)}{V^\alpha(\bx)}=\int_0^\infty g(\bs_{-t}(\bx))\left|\frac{\partial \bs_{-t}(\bx)}{\partial \bx}\right|dt,\label{integral_formula}
\end{align}
where $|\cdot|$ is the determinant. The proof follows by substituting the integral formula for $\rho(\bx)$ from (\ref{integral_formula}) in (\ref{pde}) and using the fact that 
\begin{align}\lim_{t\to \infty} g(\bs_{-t}(\bx))\left|\frac{\partial \bs_{-t}(\bx)}{\partial \bx}\right|=0.\label{limit}
\end{align} 
The limit in (\ref{limit}) goes to zero as $\rho(\bx)$ is bounded for all $\bx\in \bX_1$  and using Barbalat's Lemma. 
The integrant in (\ref{integral_formula}) defines a semi-group of linear Perron-Frobenius (P-F) operator, $\mP_t$, acting on function $g(\bx)$ and hence can be written compactly as
\begin{align}
[\mP_t g](\bx)=g(\bs_{-t}(\bx))\left|\frac{\partial \bs_{-t}(\bx)}{\partial \bx}\right|.\label{PF_def}
\end{align}
Using (\ref{PF_def}), (\ref{integral_formula}) can be written as 
\begin{align}
\rho(\bx)=\int_0^\infty [\mP_t g](\bx) dt.
\end{align}
Furthermore, (\ref{limit}) can be written as 
\[\lim_{t\to \infty}[\mP_t g](\bx)=0\implies \lim_{t\to \infty} [\mP_t \mathds{1}_{\bX_0}](\bx)=0,\]
where $\mathds{1}_{\bX_0}$ is the indicator function for set $\bX_0$. This implication follows because $g(\bx)\geq \xi>0$ for all $\bx\in \bX_0$ and from dominated convergence theorem. For any set $A\subseteq \bX_1$, we have
\begin{align}\int_A [\mP_t \mathds{1}_{\bX_0}](\bx) d\bx=\int_{\bX_1} [\mP_t \mathds{1}_{\bX_0}](\bx) \mathds{1}_{A}(\bx) d\bx\nonumber\\=\int_{\bX_1} \mathds{1}_{\bX_0} (\bx)\mathds{1}_{A}(\bs_t(\bx))dx.\label{ll}\end{align}
The above follows by using the definition of $\mP_t$ in (\ref{PF_def}) and change of variables in the integration, i.e., $\by=\bs_{-t}(\bx)$ and $d\by=|\frac{\partial \bs_{-t}(\bx)}{\partial \bx}|d\bx$ and after relabeling. Note that the right-hand side of (\ref{ll}) is nothing but
\[\int_A [\mP_t \mathds{1}_{\bX_0}](\bx) d\bx=m\{\bx\in \bX_0: \bs_t(\bx)\in A\}.\]
From Lebesgue dominated convergence theorem
\[0=\int_{A}\lim_{t\to \infty}[\mP_t\mathds{1}_{\bX_0}](\bx)d\bx\]\[=\int_{\bX_1}\mathds{1}_{\bX_0}(\bx)\lim_{t\to \infty}\mathds{1}_{A}(\bs_t(\bx))d\bx=m\{\bx\in \bX_0: \bs_t(\bx)\in A\}.\]
Since the above is true for any measurable and positive Lebesgue measure set $A\subseteq \bX_1:=\bX\setminus {\cal B}_{\delta}$ for arbitrary small $\delta$, we obtain
\begin{align}
m\{\bx\in \bX_0: \lim_{t\to \infty}\bs_t(\bx)\neq 0\}=0.
\end{align}

We next show that the unsafe set $\bX_{u_k}$ will be avoided by trajectories $\bs_t(\bx)$ starting from almost all w.r.t. Lebesgue measure initial condition $\bx\in \bX_0$. We have for $\bx\in \bX_{u_k}$
\begin{align}\rho(\bx)=\frac{\Psi_k(\bx)}{V^\alpha}=\frac{\theta}{V^\alpha}.\label{estimate}
\end{align}

Following Assumption \ref{assume_main} (equation \eqref{eq:V_min}), we have 
\begin{align}\rho(\bx)=\frac{\theta}{V^\alpha}\leq \frac{\theta}{V_{min}^k}.\label{rhobound}
\end{align}

 Using the above bound on $\rho(\bx)$, we obtain
\[\bG:=\int_{\bX_{u_k}}\int_0^\infty [\mP_t \mathds{1}_{\bX_0}](\bx)dt d\bx=\int_{\bX_{u_k}}\rho(\bx)d\bx \leq \frac{\theta}{V_{min}^k}m(\bX_{u_k}),\]
where $m(\cdot)$ is the Lebesgue measure. Utilizing that $d\by=|\frac{\partial \bs_{-t}(\bx)}{\partial \bx}|d\bx$,
which is described through the definition of 
$\mP_t$ and performing a change of variable $\by=\bs_{-t}(\bx)$, we can use the bounds on $\rho(\bx)$ in (\ref{rhobound}) for $\bx\in \bX_{u_k}$ to obtain

\[\bG=\int_{\bX_{1}} \mathds{1}_{\bX_0}(\by) \int_0^\infty  \mathds{1}_{\bX_{u_k}}(\bs_t(\by))dt d\by\leq \frac{\theta}{V_{min}^k} m(\bX_{u_k}). \]
The time integral on the left-hand side is the time spent by system trajectories starting from the initial set $\bX_0$ in the unsafe set $\bX_{u_k}$. Let this time be denoted by $T(\by)$. Hence, we obtain
\[\int_{\bX_1} T(\by) \mathds{1}_{\bX_0}(\by)d\by\leq \frac{\theta}{V_{min}^k} m(\bX_{u_k}). \]
Following Assumption \ref{assume_main} (equation \eqref{assume_epsilon}), we have
\[\theta\leq \epsilon \frac{V_{min}^k}{m(\bX_{u_k})}\implies \int_{\bX_0}T(\by)d\by\leq \epsilon, \]
for any given $\epsilon>0$.

Choose some $\eta<1$, then using  Chebyshev's inequality and the fact that $\bX_0\subset \bX_1$, we have 
\[m\{\bx\in \bX_0: T(\by)\geq \epsilon^\eta\}\leq \epsilon^{-\eta}\int_{\bX_0}T(\by)d\by\leq \epsilon^{-\eta+1}.\]

Since the above is true for arbitrary small $\epsilon>0$, we have 

\begin{align}m\{\bx\in \bX_0: T(\by)=\int_0^\infty \mathds{1}_{\bX_{u_k}}(\bs_t(\bx))dt>0\}=0.\label{ss}
\end{align}

Now we make use of  the continuity property of the flow $\bs_t(\bx)$ w.r.t. time to show that $\mathds{1}_{\bX_{u_k}}(\bs_t(\bx))=0$ for all $t\geq 0$. Assume not, then there exists  $\gamma$ and $\bar t$ such that $\mathds{1}_{\bX_{u_k}}(\bs_{\bar t}(\bx))\geq \gamma>0$. Then from the continuity of solution $\bs_t(\bx)$ w.r.t. time, we know that there exists $\Delta>0$ such that $\mathds{1}_{\bX_{u_k}}(\bs_t(\bx))>0$ for $t\in [\bar t,\bar t+\Delta]$. This violates (\ref{ss}).

\bibliographystyle{unsrt}  
\bibliography{references,Umesh_ref} 




\end{document}